\documentclass{article}

\usepackage{microtype}
\usepackage{graphicx}
\usepackage{subfigure}
\usepackage{caption}
\usepackage{booktabs} 

\usepackage{algorithm}
\usepackage{algorithmic}

\usepackage{lipsum}
\usepackage{tikz}
\usetikzlibrary{shapes.geometric, arrows}

\usepackage[pagebackref,breaklinks,colorlinks]{hyperref}

\usepackage[accepted]{icml2023}

\usepackage{amsmath}
\usepackage{amssymb}
\usepackage{mathtools}
\usepackage{amsthm}

\usepackage[capitalize,noabbrev]{cleveref}

\theoremstyle{plain}

\theoremstyle{definition}

\theoremstyle{remark}

\usepackage[textsize=tiny]{todonotes}

\tikzstyle{startstop} = [rectangle, rounded corners, minimum width=3cm, minimum height=1cm,text centered, draw=black, fill=red!30]
\tikzstyle{startstop2} = [rectangle, rounded corners, minimum width=3cm, minimum height=1cm,text centered, draw=black, fill=blue!30]
\tikzstyle{startstop3} = [rectangle, rounded corners, minimum width=3cm, minimum height=1cm,text centered, draw=black, fill=green!30]
\tikzstyle{startstop4} = [rectangle, rounded corners, minimum width=3cm, minimum height=1cm,text centered, draw=black, fill=orange!30]
\tikzstyle{process} = [rectangle, minimum width=3cm, minimum height=1cm, text centered, draw=black, fill=orange!30]
\tikzstyle{decision} = [diamond, minimum width=3cm, minimum height=1cm, text centered, draw=black, fill=green!30]
\tikzstyle{arrow} = [thick,->,>=stealth]
\tikzstyle{dashedarrow} = [thick, dashed,->,>=stealth]

\begin{document}

\twocolumn[

\icmltitle{Learning by Grouping: A Multilevel Optimization Framework for Improving Fairness in Classification without Losing Accuracy}





\icmlsetsymbol{equal}{*}

\begin{icmlauthorlist}
\icmlauthor{Ramtin Hosseini}{yyy}
\icmlauthor{Li Zhang}{yyy}
\icmlauthor{Bhanu Garg}{yyy}
\icmlauthor{Pengtao Xie}{yyy}
\end{icmlauthorlist}

\icmlaffiliation{yyy}{Department of Electrical and Computer Engineering, UCSD, San Diego, USA}

\icmlcorrespondingauthor{Ramtin Hosseini}{rhossein@eng.ucsd.edu}

\icmlkeywords{Machine Learning, ICML}

\vskip 0.3in
]



\printAffiliationsAndNotice{}  

\begin{abstract}

The integration of machine learning models in various real-world applications is becoming more prevalent to assist humans in their daily decision-making tasks as a result of recent advancements in this field. However, it has been discovered that there is a tradeoff between the accuracy and fairness of these decision-making tasks. In some cases, these AI systems can be unfair by exhibiting bias or discrimination against certain social groups, which can have severe consequences in real life. Inspired by one of the most well-known human learning skills called \textit{grouping}, we address this issue by proposing a novel machine learning (ML) framework where the ML model learns to group a diverse set of problems into distinct subgroups to solve each subgroup using its specific sub-model. Our proposed framework involves three stages of learning, which are formulated as a three-level optimization problem: 1) learning to group problems into different subgroups; 2) learning group-specific sub-models for problem-solving; 3) updating group assignments of training examples by minimizing the validation loss. These three learning stages are performed end-to-end in a joint manner using gradient descent. To improve fairness and accuracy, we develop an efficient optimization algorithm to solve this three-level optimization problem. To further decrease the risk of overfitting in small datasets using our LBG method, we incorporate domain adaptation techniques in the second stage of training. We further apply our method to differentiable neural architecture search (NAS) methods. Extensive experiments on CIFAR-10, CIFAR-100, ImageNet, ISIC-18, and CelebA datasets demonstrate our methods’ effectiveness and performance improvements in both fairness and accuracy. Our proposed \textit{Learning by Grouping} (LBG) can reduce overfitting and achieve state-of-the-art performances with fixed human-designed network architectures and searchable network architectures on various datasets.

\end{abstract}

\section{Introduction}
\label{sec:1}

\begin{figure}
    \centering
    \includegraphics[width=0.5\textwidth]{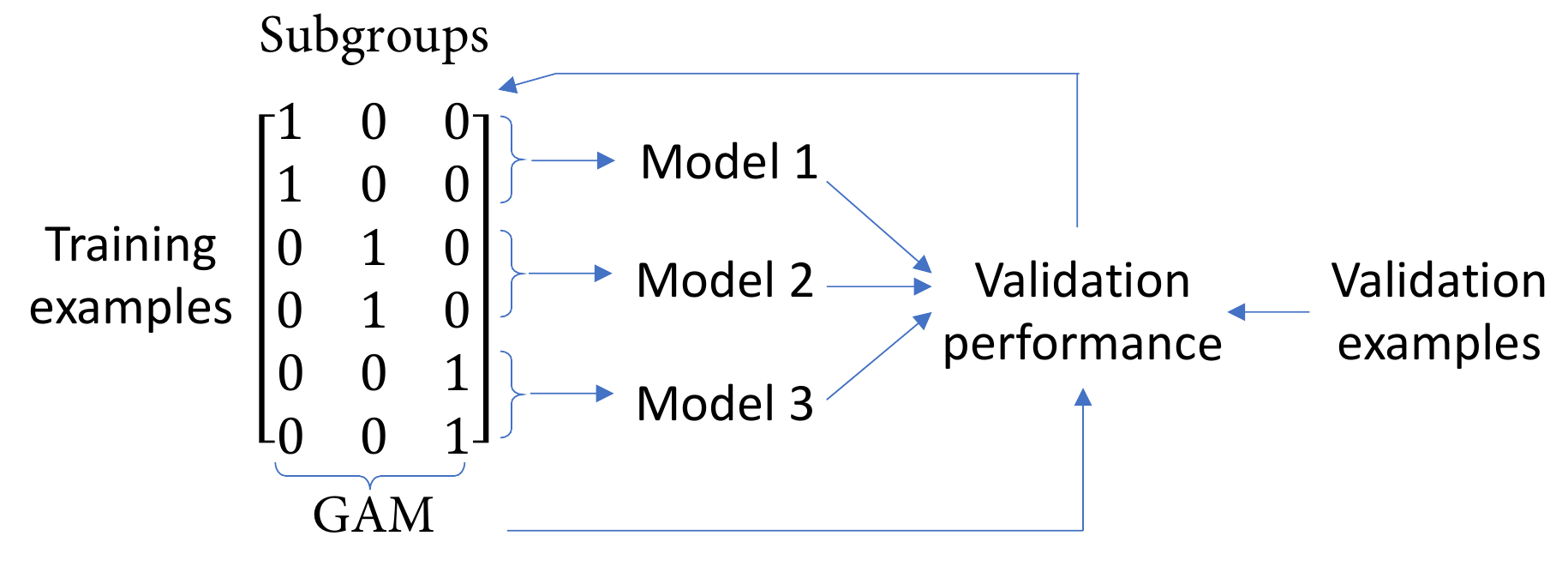}
    \caption{Illustration of Learning by Grouping (LBG) with three subgroups (i.e., $K=3$). As shown, we update our \textit{Group Assignment Model} (GAM) for the training examples by validating the performance of the validation set, which contrasts LBG method with the existing MoE approaches.
    }
    \label{fig:1}
\end{figure}

Learning by grouping is an outstanding human learning skill aiming to organize a set of given problems into different subgroups and domains where each subgroup contains similar problems that can be solved independently and efficiently. In this paper, we formulate \textit{Learning by Grouping} (LBG) as an optimization problem and investigate its effectiveness in ML. Our proposed framework contains two types of model: 1) Group Assignment Model (GAM); and 2) Group-Specific Classification Models (GSCM). The GAM model takes a data example as input and predicts the subgroup it belongs to -- a $K$-way classification problem, where $K$ is the number of GSCM models  (i.e., experts). For each subgroup $k$, a  GSCM model performs the supervised learning on the target task. We then apply the GAM and the $K$ GSCM models to improve the existing machine learning models' fairness and accuracy. Additionally, we extend our LBG formulation to the neural architecture search to obtain the most suitable task-specific GSCM models. We depict the high-level learning process in Fig \ref{fig:1}.

We formulate LBG as a three-stage optimization problem. First, we learn the Group-Assignment Model (GAM); then, we train Group-Specific Classification Models (GSCMs); finally, we apply the GAM and the GSCMs to the validation set to learn the subgroups for subgroup assignment and the learnable architecture. We develop a gradient-based method to solve this three-level optimization problem. In previous related works, mixture-of-expert methods learn the experts -- analogous to GSCMs; and the gating network -- similar to GAM. The mixture-of-experts (MoE) methods learn the gating network and the experts jointly on the training data, which has a high risk of overfitting the gating network to the training data. We address this problem of overfitting by MoE methods by formulating a three-stage optimization framework that learns the subgroups for the subgroup assignment tasks on the validation set instead of the training examples.  

Currently, the majority of state-of-the-art neural network performance is achieved through architectures that are manually designed by humans. However, this process of designing and evaluating neural network architectures by human experts is both time-consuming and may not end with the most suitable task-specific architecture. In recent years, there has been a growing interest in automating this manual process, referred to as neural architecture search (NAS). On the other hand, humans possess powerful learning skills that have been developed through evolution. This study also examines the potential of using a human-based learning technique, known as learning by grouping, in differentiable NAS approaches.

The major contributions of this paper include:
\begin{itemize}
    \item Drawing inspirations from the human learning technique of \textit{Learning by Grouping} (LBG), we propose a new machine learning framework that utilizes grouping to divide a set of diverse problems into distinct subgroups. The proposed framework groups similar problems together within each subgroup and subsequently develops a group-specific solution for each subgroup.
    \item We propose a three-level optimization framework to formulate LBG. We provide a solution to solve the optimization problem jointly end-to-end via gradient descent: 1) learning to group problems into different subgroups; 2) learning group-specific sub-models; 3) learning group-assignments of training examples by minimizing the validation loss. 
    \item We also propose \textit{domain adaptive} LBG (DALBG) to mitigate the risk of overfitting within our LBG framework by utilizing domain adaption techniques.
    \item We extend the above formulation to the challenging neural architecture search (NAS) problem, and we show that LBG/DALBG can be applied to any differentiable NAS approach for further improvements.
    \item We perform experiments on CelebA, ISIC-18, CIFAR-10, CIFAR-100, and ImageNet datasets to showcase the effectiveness of our proposed method in both fairness and accuracy aspects. Additionally, we apply our proposed LBG to language understanding tasks by conducting experiments on GLUE datasets, which can be found in the Supplements.
    
\end{itemize}

\section{Related Works}

\subsection{Mixture of Experts}
\vspace{-0.1 cm}
Lately, a wide variety of works \cite{shazeer2017outrageously, DBLP:conf/iccv/ZhangHLT19, wang2020deep} have proposed applying the mixture-of-experts (MoE) approach, which was initially proposed by \cite{JacobsJordanNowlanEtAl91}, to varied deep learning tasks. Generally, deep learning MoE frameworks consist of expert networks and a gating function, where the gating function assigns each expert a subset of training data. The methods assume a set of latent experts where each expert performs a classification or regression task. A gating function assigns the given data example to an expert. Then this example is classified using the classification model specific to this expert. The MoE has been an active research area aiming to improve the vanilla ML approaches, such as \cite{shazeer2017outrageously, DBLP:conf/iccv/ZhangHLT19, wang2020deep}. \cite{shazeer2017outrageously} introduces a trainable gating function to assign the experts' sparse combinations for the given data. DeepMOE~\cite{wang2020deep} proposes a deep convolutional network including a shallow embedding network and a multi-headed sparse gating network, where the multi-headed sparse gating network uses the mixture weights computed by the shallow embedding network to select and re-weight gates in each layer. In MGE-CNN\cite{DBLP:conf/iccv/ZhangHLT19}, experts are learned with the extra knowledge of their previous experts along with a Kullback-Leibler (KL) divergence constraint to improve the diversity of the experts. Recently, \cite{riquelme2021scaling} proposed the Vision Transformer MoE (V-MoE) that can successfully reach state-of-the-art on ImageNet with approximately half of the required resources.

In the existing MoE methods, which are based on single-level optimization, the gating function and expert-specific Classification Models are learned jointly by minimizing the training loss. Hence, there is a high risk of the gating function overfitting the training data, which can lead to unfair and inaccurate decision-making. In our method, we address this issue via learning the group assignments of training examples by minimizing the validation loss instead and developing a multi-stage optimization problem rather than joint training. The results show the efficacy of our method. 

\vspace{-0.1 cm}
\subsection{Domain Adaptation}
\vspace{-0.1 cm}
Domain adaptation (DA) is a technique in machine learning that aims to enhance the performance of models trained on one domain, known as the source domain, on a different yet related domain, referred to as the target domain. The objective is to transfer the knowledge acquired from the source domain to the target domain, where the input features and/or output labels may vary. This approach is particularly valuable in scenarios where the amount of labeled data in the target domain is scarce, but a large amount of labeled data is available in the source domain. 
Different methods for domain adaptation \cite{Gretton2009CovariateSB,6126344,5640675,6247924} can be classified into three main categories: instance-based, feature-based, and adversarial-based approaches. These methods mostly focus on measuring and minimizing the distance between the source and target domains. Some well-established distance measuring approaches include Maximum Mean Discrepancy (MMD)~\cite{long2015learning,gretton2008kernel}, Correlation Alignment (CORAL)~\cite{sun2017correlation}, Kullback-Leibler (KL) divergence~\cite{10.2307/2236703}, and Contrastive Domain Discrepancy (CDD)~\cite{kang2019contrastive}.

\subsection{Multi-Level Optimization}
In the past few years, Bi-Level Optimization (BLO) and Multi-Level Optimization (MLO) \cite{Vicente94bileveland} techniques have been applied to Meta-Learning~\cite{10.5555/2887007.2887164,finn2017model}, and Automated Machine Learning (AutoML) tasks such as neural architecture search \cite{cai2018proxylessnas,liu2018darts,xie2018snas,abs-1907-05737,liang2019darts+,hosseini2021dsrna} and hyperparameter optimization \cite{10.5555/2887007.2887164,baydin2017online} to learn the meta parameters automatically and reduce the required resources and reliance on humans for designing such methods. Lately, inspired by humans' learning skills \cite{xie2020skillearn}, several existing works \cite{10.1145/3503161.3548150,chitnis2022brain,hosseini2020learning,Garg2021LearningFM,du2020learning,hosseini2022saliencyaware,sheth2021learning,du2020small,zhu2022self} have borrowed these skills from humans and extended them to ML problems in MLO frameworks to study whether these techniques can assist the ML models in learning better.

\subsection{Neural Architecture Search} 
Recently, Neural Architecture Search (NAS) has attracted the researchers' attention to assist them in finding high-performance neural architectures for different deep learning applications. In the early stages, most of the proposed NAS methods were based on reinforcement learning (RL)  \cite{zoph2016neural,pham2018efficient,zoph2018learning} and evolutionary learning  \cite{liu2017hierarchical,real_regularized_2019}. Reinforcement learning approaches use a policy network to generate architectures by maximizing the accuracy of the validation set, which is used as a reward. In evolutionary learning methods, architectures describe the individuals of a population, and the validation accuracy of the individuals is used as fitness scores. Replacing low fitness scores individuals with higher fitness scores individuals leads to enhanced performance. Reinforcement learning and evolutionary learning approaches are computationally expensive. To solve for the high computational cost by the RL and evolutionary learning-based methods, the research community introduced differentiable search methods \cite{cai2018proxylessnas,liu2018darts,xie2018snas}, which are extremely efficient compared to the previous methods since they use the weight-sharing techniques and perform the searching process using gradient descent. Differentiable NAS was first proposed by DARTS~\cite{liu2018darts}. Lately, many following works \cite{chen2019progressive,abs-1907-05737,liang2019darts+} have worked on enhancing the search results and reducing the computational cost of differentiable NAS even further. For instance, P-DARTS~\cite{chen2019progressive} increases the depth of architectures progressively during the search. PC-DARTS~\cite{abs-1907-05737} reduces the redundancy by evaluating only a subset of channels in the search process. 

\section{Methods}
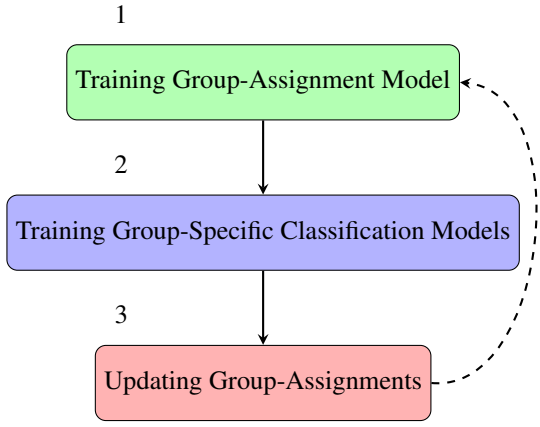
\begin{figure}
	\centering
\begin{tikzpicture}[node distance=2cm]
\node (start) [startstop3] {Training Group-Assignment Model};
\node (num1) [above right of=start, xshift=-3.3cm, yshift=-0.5cm] {1};
\node (sub1) [startstop2, below of=start] {Training Group-Specific Classification Models};
\node (num2) [above right of=sub1, xshift=-3.3cm, yshift=-0.5cm] {2};
\node (sub2) [startstop, below of=sub1] {Updating Group-Assignments};
\node (num2) [above right of=sub2, xshift=-3.3cm, yshift=-0.5cm] {3};

\draw [arrow] (start) -- (sub1);
\draw [arrow] (sub1) -- (sub2);
\draw [dashedarrow] (sub2) to [bend right=90] (start);

\end{tikzpicture}
\caption{Overview of our proposed three-level optimization framework (Learning by Grouping).}
\label{fig:flowchart}
\end{figure}

Our method consists of a Group-Assignment Model (GAM) and $K$ Group-Specific Classification Models (GSCMs). The GAM model predicts and assigns the training samples to their corresponding GSCM expert model. Then the GSCM models predict the classes of the inputs. Lastly, we apply the GAM and the GSCMs to the validation set and minimize the validation loss to learn the assignments of training samples. The illustration of our proposed method is shown in Fig \ref{fig:flowchart}.
In Section \ref{sec:3.1}, we first begin with defining the three-level optimization framework to formulate LBG (Section \ref{sec:3.1.1}), and then we integrate domain adaptation techniques to our proposed LBG to mitigate the risk of overfitting (Section \ref{sec:3.1.2}). Afterward, we extend the LBG to the Neural Architecture Search problem in Section \ref{sec:3.1.3}. Finally, in Section \ref{sec:3.2}, we develop an efficient optimization algorithm to address the three-level optimization problem.

\subsection{Three-Level Optimization Framework}
\label{sec:3.1}
Our framework is composed of two types of models: the Group-Assignment Model (GAM) and the Group-Specific Classification Models (GSCM). The GAM model takes a data example as input and assigns it to one of the subgroups, which is a K-way classification problem, where K is the number of GSCM models (i.e., experts). We propose an end-to-end three-stage optimization problem where: First, the Group-Assignment Model (GAM) is learned; then, the Group-Specific Classification Models (GSCMs) are trained; finally, the GAM and the GSCMs are applied to the validation set to determine the group assignments and the learnable architecture. As shown in Fig \ref{fig:1}, the Group-Assignment Model (GAM) is updated for training examples by validating the performance on the validation set, which distinguishes LBG from existing MoE approaches. As discussed in Section \ref{sec:3.1.1}, the group assignments from GAM are continuous values $C_{nk}\in [0,1]$.
Therefore, to convert these probability distributions to one-hot encoded format (similar to Fig \ref{fig:1}) we can compute the top-k  and obtain the k-hot encoded matrix, where $k$ is one in this case.


\subsubsection{Learning by Grouping (LBG)}
\label{sec:3.1.1}
We assume there are $K$ latent subgroups. Let $C$ be a  matrix denoting the learnable `ground-truth' grouping of the training samples. The size of $C$ is $N \times K$ where $N$ is the number of training examples - row $n$ represents the grouping of the $n$-th training example. We relax the values in each row from a one-hot encoding to continuous values in order to perform gradient descent, so that $C_{nk}\in [0,1]$ denotes the probability that the $n$-th training example belongs to the $k$-th latent subgroup. Subgroups $C$ are initialized randomly. The latent subgroup labels for subgroups are permutation-invariant.  We then assign the $n$-th training example to subgroup $j_n$ such that $j_n = \arg \max_e C_{ne}$, and let $G_{n} = C_{nj_n}$ be the probability of grouping the sample $x_n$ to subgroup $j_n$. Let the GAM be represented by $f(x_n;T)$ with SoftMax output, which takes a data example $x_n$ as input and predicts which subgroup $x_n$ should be assigned to. $T$ is the weights parameter of this network. The output of $f(x_n;T)$ is a $K$-dimensional vector, where the $k$-th element $f_k(x_n;T)$ denotes the probability that $x_n$ should be assigned to the $k$-th subgroup. The sum of elements in $f(x_n;T)$ is one. Let $\hat{j}_n = \arg \max_e f_{e}(x_n;T)$, and let $E_{n} = f_{\hat{j}_n}(x_n;T)$ be the confidence of the GAM in assigning $x_n$ the subgroup $\hat{j}_n$. We then have a GSCM  classifier $f(x_n;S_{\hat{j}_n})$ for each latent subgroup $\hat{j}_n \in \{1 \dots K\}$, which predicts the class label for a data example $x_n$ that has been assigned $\hat{j}_n$ as its GSCM  by the GAM: $f(x_n;T)$. $S_{\hat{j}_n}$ are the network weights of this GSCM  classifier. 

\paragraph{Stage I.} In the first stage, we optimize the GAM: $f(x;T)$ given $C$ by solving the following `relaxed' negative log-likelihood optimization problem:
\begin{equation}\label{stage1}
\resizebox{.7\columnwidth}{!}{
    $T^*(C)=\underset{T}{\text{argmin}} \; \sum_{n=1}^N  -G_n(C)\log f_{j_n}(x_n;T)$ 
    }
\end{equation}
Note that we do not update the `ground-truth' subgroups $C$ in this stage.

\paragraph{Stage II.} In the second stage, we learn the $K$ GSCM models. 
For each latent subgroup $k$, there is a GSCM classifier $f(x;S_k)$ with parameters $S_k$, which predicts the class label for a data example $x$ assigned $k$ as its subgroup by the GAM. Let $D_n=\{x_{n}\}$ denote the  subset of training data examples assigned subgroup $k$ by GAM. We want to learn $S_k$ by minimizing the following loss:
\begin{equation}
\resizebox{.4 \columnwidth}{!}{
    $\sum_{x_n\in D_n} \ell(f(x_n;S_k),y_n)$
    }
\end{equation}
where $y_n$ is the class label of $x_n$ . $l(\cdot,\cdot)$ is the cross-entropy loss. In addition, we take into account the confidence of the GAM in assigning the training example $x_n$ to its corresponding GSCM $\hat{j}_n$. So we relax the above equation, and summarize the total loss of all GSCM models and objective of this stage as:
\begin{equation}\label{stage2}
\resizebox{.9 \columnwidth}{!}{
   $\{S^*_k(T^*(C))\}_{k=1}^K=
     \underset{\{S_k\}_{k=1}^K}{\text{argmin}} \sum_{n=1}^N E_{n}\left(x_n;T^*(C)\right) \ell(f(x_n;S_{\hat{j}_n}),y_n)$
     }
\end{equation}
where  $S^*_k(T^*(C))$ for $k \in \{1 \dots K\}$ denotes the optimal solution set for the $K$ GSCM  classifiers. 

\paragraph{Stage III.} Given $T^*(C)$ and $S^*_k(T^*(C))$, we apply them to make predictions on the validation examples and update the  `ground-truth' matrix $C$. The validation loss is:
\begin{equation}\label{stage3}
\resizebox{.8 \columnwidth}{!}{
    $\underset{C}{\text{min}} \sum_{i=1}^M \ell\left( E_{\hat{j}_i}\left(x_i;T^*(C)\right)
    f\left(x_i; S^*_{\hat{j}_i}(T^*(C))\right), y_i\right)$
    }
\end{equation}
where $\hat{j}_i = \arg \max_e f_e(x_i;T^*(C)) $ and $M$ is the number of validation examples. $y_i$ is the class label of $x_i$. We update $C$ by minimizing this validation loss. 

Putting these pieces together, we have the following optimization problem:
\begin{equation}
\resizebox{.75\columnwidth}{!}{
    $\underset{C}{\min}    \sum_{i=1}^M \ell\left( E_{\hat{j}_i}\left(x_i;T^*(C)\right)
    f\left(x_i; S^*_{\hat{j}_i}(T^*(C))\right), y_i\right) \nonumber $
}
\end{equation}
\begin{equation}\label{fixOP}
\resizebox{.9\columnwidth}{!}{
   $ s.t.   \{S^*_k(T^*(C))\}_{k=1}^K= 
     \underset{\{S_k\}_{k=1}^K}{\text{argmin}} \sum_{n=1}^N E_{n}\left(x_n;T^*(C)\right) \ell(f(x_n;S_{\hat{j}_n}),y_n)$
     }
\end{equation}
\begin{equation}
\resizebox{.6\columnwidth}{!}{
     $T^*(C)=\underset{T}{\text{argmin}} \sum_{n=1}^N  -G_n(C)\log f_{j_n}(x_n;T)  \nonumber $
     }
\end{equation}



\subsubsection{Domain Adaptive LBG}
\label{sec:3.1.2}

In our proposed \textit{Learing by Grouping} (LBG) from Section \ref{sec:3.1.1}, the $N$ training examples are divided into $K$ subgroups. As a result, each subgroup has approximately $N/K$ training examples. The reduced number of training examples in small datasets can potentially lead to higher risk of overfitting for each subgroup. To address this problem, we propose domain-adaptive LBG (DALBG) where we treat each subgroup as a domain. During the second stage of our framework, when we are training a group-specific classifier for a subgroup $k$, we perform domain adaptation to adapt examples from other subgroups into subgroup $k$ and use these adapted examples as additional training data for subgroup $k$. 
For the sake of simplicity, our proposed framework employs the MMD-based \cite{long2015learning} domain adaptation approach. However, it should be noted that 
other domain adaptation techniques can also be incorporated within our framework.
For a specific subgroup, $k$, let $\{x_i^k\}_{i=k}^{N_k}$ represent the examples assigned to this subgroup and $\{x_j^{-k}\}_{j=1}^{N-N_k}$ represent the examples not assigned to this subgroup. In order to adapt $\{x_j^{-k}\}_{j=1}^{N-N_k}$ into subgroup $k$, we minimize the Maximum Mean Discrepancy (MMD) loss as follows:
\begin{equation}
\resizebox{.7\columnwidth}{!}{
$M_{k}=\left\|\frac{1}{N_{k}} \sum_{i=1}^{N_{k}} \phi\left(x_{i}^{k} ; S_{k}\right)-\frac{1}{N-N_{k}} \sum_{j=1}^{N-N_{k}} \phi\left(x_{j}^{-k} ; S_{k}\right)\right\|_{2}^{2}$
}
\end{equation}
where $\phi\left(x_{i}^{k} ; S_{k}\right)$ denotes the embedding of $x_{i}^{k}$ extracted by $S_{k}$. This loss can be relaxed to:
\begin{equation}\label{second2DA}
\resizebox{.9\columnwidth}{!}{
$M_{k}=\left\|\frac{1}{N} \sum_{n=1}^{N} f_{k}\left(x_{n} ; T^{*}(A)\right) \phi\left(x_{n} ; S_{k}\right)-\frac{1}{N} \sum_{n=1}^{N}\left(1-f_{k}\left(x_{n} ; T^{*}(A)\right)\right) \phi\left(x_{n} ; S_{k}\right)\right\|_{2}^{2}$
}
\end{equation}

Thus, by adding $M_{k}$ to our second stage Eq. (\ref{stage2}) we define the following \textit{domain adaptive} LBG (DALBG) problem:
\begin{equation}
\resizebox{.7\columnwidth}{!}{
    $\underset{C}{\min}    \sum_{i=1}^M \ell\left( E_{\hat{j}_i}\left(x_i;T^*(C)\right)
    f\left(x_i; S^*_{\hat{j}_i}(T^*(C))\right), y_i\right) \nonumber $
}
\end{equation}
\begin{equation}\label{DA}
\resizebox{.91\columnwidth}{!}{
   $ s.t.   \{S^*_k(T^*(C))\}_{k=1}^K= 
     \underset{\{S_k\}_{k=1}^K}{\text{argmin}} \sum_{n=1}^N E_{n}\left(x_n;T^*(C)\right) \ell(f(x_n;S_{\hat{j}_n}),y_n)+\lambda M_{k} $
     }
\end{equation}
\begin{equation}
\resizebox{.55\columnwidth}{!}{
   $ T^*(C)=\underset{T}{\text{argmin}} \sum_{n=1}^N  -G_n(C)\log f_{j_n}(x_n;T)  \nonumber $
     }
\end{equation}
where $\lambda$ is a tradeoff parameter. Note that our proposed LBG in Eq. (\ref{fixOP}) method is a special case of DALBG in Eq. (\ref{DA}) with $\lambda=0$. For the sake of simplicity, we refer to both Learning by Grouping \textit{\textbf{with/without}} domain adaptation as (DA)LBG.

\subsubsection{Neural Architecture Search Application}

\label{sec:3.1.3}
In this section, we extend the formulation in Eq. (\ref{fixOP}) to be applicable to neural architecture search. Similar to~\cite{liu2018darts}, the $k$-th GSCM has a differentiable architecture $A_k$. The search space of $A_k$ is composed of large number of building blocks, where the output of each block is associated with a weight $a$ indicating the importance of the block. After learning, the block whose weight $a$ is among the largest are retained to form the final architecture. To this end, architecture search amounts to optimizing the set of architecture weights $A_k = \{a\}$. 


\textbf{Stage I} and \textbf{Stage II} have the same procedure as Eq. (\ref{stage1}) and Eq. (\ref{second2DA}). In the second stage, the network weights $S_k$ of the expert model are a function of its architecture $A_k$. We keep the architecture fixed at this stage, and learn the weights $S_k(A_k)$. However, \textbf{Stage III} does not precisely follow Eq. (\ref{stage3}). Given $T^*(C)$ and $S^*_k(A_k,T^*(C))$, we apply them to make predictions on the validation examples and update the  `ground-truth' matrix $C$, as well as the architectures $A_k$ based on the validation loss, where $k \in \{1 \dots K\}$ . Hence, we update Eq. (\ref{stage3}) as follows: 
\begin{equation}
\resizebox{.8 \columnwidth}{!}{
    $\underset{C, \{A_k\}_{k=1}^K}{\min}   \sum_{i=1}^M \ell\left( E_{\hat{j}_i}\left(x_i;T^*(C)\right)
    f\left(x_i; S^*_{\hat{j}_i}(A_{\hat{j}_i}, T^*(C))\right), y_i\right)$
    }
\end{equation}

Thus, the overall optimization problem with learnable architecture is as follows:
\begin{equation}
    \resizebox{.75 \columnwidth}{!}{
  $\underset{C, \{A_k\}_{k=1}^K}{\min}     \sum_{i=1}^M \ell\left( E_{\hat{j}_i}\left(x_i;T^*(C)\right)
    f\left(x_i; S^*_{\hat{j}_i}(A_{\hat{j}_i}, T^*(C))\right), y_i\right) \nonumber $
    }
 \end{equation}   
 \begin{equation}
    \label{NASOP}
    \resizebox{.89 \columnwidth}{!}{
    $  s.t.   \{S^*_k(A_k, T^*(C))\}_{k=1}^K=
     \underset{\{S_k\}_{k=1}^K}{\text{argmin}} \sum_{n=1}^N E_{n}\left(x_n;T^*(C)\right) \ell(f(x_n;S_{\hat{j}_n}(A_{\hat{j}_n}),y_n)+\lambda M_{k}$
     }
     \end{equation}
     \begin{equation}
     \resizebox{.55 \columnwidth}{!}{
      $ T^*(C)=\underset{T}{\text{argmin}}  \sum_{n=1}^N  -G_n(C)\log f_{j_n}(x_n;T)  \nonumber $
       }
\end{equation}

 Our framework is orthogonal to existing differentiable NAS methods, and hence can be applied on top of any like  DARTS~\cite{liu2018darts}, P-DARTS~\cite{chen2019progressive}, PC-DARTS~\cite{abs-1907-05737}, and DARTS$-$~\cite{abs-2009-01027} among the others.

\subsection{Optimization Algorithm}
\label{sec:3.2}
We promote an efficient algorithm to solve the LBG, DALBG, and LBG-NAS problems described in Eq.~(\ref{fixOP}), Eq.~(\ref{DA}), and Eq.~(\ref{NASOP}), respectively. We utilize a fairly similar procedure as~\cite{liu2018darts}  to calculate the gradient of Eq.~(\ref{stage1}) w.r.t $T$ and approximately update $T^*(C)$ via one-step gradient descent. Then since DALBG in Eq.~(\ref{DA}) is the generalized version of LBG in Eq.~(\ref{fixOP}), we plug the approximation $T^{'}(C)$ into the Eq.~(\ref{second2DA}) to get an $O_{S_k}$, which denotes the approximated objective of  $S_k$. Similarly to the previous step, we approximate $S^*_k(T^{'}(C))$ using a one-step gradient descent update of $S_k$ based on the gradient of the approximated objective. Note that in LBG-NAS, we approximate $S^*_k(A_k, T^*(C))$, which is also a function of architecture $A_k$. Finally, we plug the approximations $T^{'}(C)$ and  $S^{'}_{k}(T^{'}(C))$ into the third stage equations to get the third approximate objective denoted by $O_C$. $C$ can be updated using  gradient descent on $O_C$. In LBG-NAS, we update the architectures $\{A_k\}^K_{k=1}$, as well. Thus, use the same approach to find the approximate objective of the architectures $\{A_k\}$ : $O_{\{A_k\}}$ for each $k \in \{1\dots K\}$, and we update it using gradient descent. We do these steps until convergence. 
\section{Experiments}
\label{sec:4}
In this section, we investigate the effectiveness of our proposed (DA)LBG framework with both fixed human-designed GSCMs and searchable GSCMs. The differentiable NAS approach consists of architecture search and evaluation stages, where the optimal cell obtained from the search stage is stacked several times into a larger composite network. We then train the resultant composite network from scratch in the evaluation stage. Please refer to the appendix (supplements) for information on adapting our method for language understanding tasks.


    
    

\subsection{Datasets}
Various experiments are conducted on four datasets: ISIC-18, CelebA, CIFAR-10, CIFAR-100, and ImageNet~\cite{deng2009imagenet} for image classification.
The CelebA dataset, consisting of 200k images of human faces with 40 features per image \cite{liu2015deep}, is used in this study. From the dataset, we select a sample of 10,000 images, with 70\% allocated for training, 15\% for validation, and 15\% for testing. 
The Skin ISIC 2018 dataset \cite{codella2019skin,tschandl2018ham10000} comprises a total of 10,015 dermatological images, intended for the purpose of 7-class skin cancer classification. In this paper, we choose gender (male and female) as our sensitive attribute that may harbor bias. To address this potential source of bias, we have randomly partitioned the dataset into training and testing sets, with the training set accounting for 80\% and the testing set accounting for 20\% of the overall dataset.
CIFAR-10 contains of 10 classes and CIFAR-100 contains  100 classes. Each dataset holds 60K images. For each of the datasets, during grouping and architecture search processes, we use 25K images as the training set, 25K images as the validation set, and the rest of the 10K images as the test set. During grouping and architecture evaluations, the combination of the above training and validation set is used as the training set of size 50k images.  ImageNet carries 1.2M training images and 50K test images with 1000 classes. Due to extensive amount of images in ImageNet, the architecture search can be pretty costly. Thus, following~\cite{abs-1907-05737}, we randomly choose 10\%, and 2.5\% of the 1.2M images to create a new training set and validation set, respectively, for the architecture search phase.  Then, we utilize all the 1.2M images through the evaluation.

\subsection{Experimental Settings}
\label{sec:4.2}
We compare the (DA)LBG image classification tasks with fixed architectures to the following MoE baselines: ResNet \cite{he2016deep}, Swin-T \cite{liu2021swin}, T2T-ViT \cite{yuan2021tokens}, DeepMOE \cite{wang2020deep}, and MGE-CNN \cite{DBLP:conf/iccv/ZhangHLT19}. Next, we compare LBG-NAS on image classification with DARTS-based methods including DARTS~\cite{liu2018darts}, P-DARTS~\cite{chen2019progressive}, and PC-DARTS~\cite{abs-1907-05737}. To ensure the training costs of our methods  with $K$ GSCM models are similar to those of baselines, we reduce the parameter number of each expert to $1/K$ of the parameter number of the baseline models by reducing the number of layers in each GSCM model. In this way, the total size of our methods are comparable to the baselines. In addition, we train each group-specific sub-model only using examples assigned to its corresponding subgroup, rather than using all training examples. So the computation cost is $O(N)$ rather than $O(NK)$, where $N$ is the number of training examples and $K$ is the number of latent subgroups. In each iteration of the algorithm, we use minibatches of training examples to update sub-models, which further reduces computation cost.

\begin{center}
\begin{table}
\centering
\caption{Results on CelebA with the target label of "attractive" and sensitive attribute of "gender".}
\vspace{.2cm}
\resizebox{.99\columnwidth}{!}{
    \centering
    \begin{tabular}{l|cccc}
  \hline
  Methods &  Error (\%) & DP &DEO & Architecture \\

       \hline
       \hline
       ResNet18  &  17.57 & 0.5023 & 0.5683 & Manual\\
       LBG-ResNet18   (ours)&  \underline{17.02} & \underline{0.2173} & \textbf{0.0596} & Manual\\
       DALBG-ResNet18 (ours)&  \textbf{16.84} &  \textbf{0.2116} & \underline{0.0835} & Manual\\
       
       \hline
       
       DARTS  & 16.39 & 0.4571 & 0.3606& NAS\\
       LBG-DARTS (ours) & \underline{15.91}  & \textbf{0.2149} & \textbf{0.0535} & NAS \\
       DALBG-DARTS (ours) & \textbf{15.22}  & \underline{0.2185} & \underline{0.0891}& NAS \\
       \hline
    \end{tabular}
    }
    \label{tab:fair}  
\end{table}
\end{center}


    

\begin{center}
\begin{table}[ht]
\centering
\vspace{-1cm}
\caption{Results of ISIC when the sensitive attribute is "gender".}
\vspace{.2cm}
\resizebox{.99\columnwidth}{!}{
    \centering
    \begin{tabular}{l|cccc}
    \hline
    Methods & Error(\%) & SPD & EOD & AOD  \\
    \hline
    \hline
    ResNet18  & 14.3 & 0.114 & 0.143 & 0.170 \\
    LBG-ResNet18 (ours)  & \textbf{12.8} & \textbf{0.051} & \textbf{0.088}& \textbf{0.074}\\
    \hline
    \hline
    DARTS & 10.2 & 0.121 & 0.139 & 0.154 \\
    LBG-DARTS (ours) & \textbf{8.4} & \textbf{0.048}  & \textbf{0.065} & \textbf{0.069}\\
    \hline
    \end{tabular}
    }
    
    \label{tab:isic-fair}
\end{table}
\vspace{-1cm}
\end{center}


\begin{table}
    \centering
    \caption{Test errors comparison of vanilla (base) models, baselines and LBG on CIFAR-10, CIFAR-100, and ImageNet.}
    \begin{tabular}{l||c|c}
    \hline
    Dataset & Model &Error(\%)\\
    \hline
    CIFAR-10&ResNet56 (vanilla)& 6.55\\
    CIFAR-10&DeepMOE-ResNet56& 6.03\\
    CIFAR-10&MGE-CNN-ResNet56& 5.91\\
    CIFAR-10&LBG-ResNet56 (ours)&5.53\\
    CIFAR-10&DALBG-ResNet56 (ours)&\textbf{5.47}\\
    \hline
    \hline
    CIFAR-100&ResNet56 (vanilla)& 31.46\\
    CIFAR-100&DeepMOE-ResNet56& 29.77\\
    CIFAR-100&MGE-CNN-ResNet56& 29.82\\
    CIFAR-100&LBG-ResNet56 (ours)&27.96\\
    CIFAR-100&DALBG-ResNet56 (ours)&\textbf{27.95}\\
    \hline
    \hline
    ImageNet&ResNet18 (vanilla)&30.24\\
    ImageNet&DeepMOE-ResNet18&29.05\\
    ImageNet&MGE-CNN-ResNet18&29.30\\
    ImageNet&LBG-ResNet18 (ours)&28.21\\
    ImageNet&DALBG-ResNet18 (ours)&\textbf{28.08}\\
    \cline{2-3}
    ImageNet& T2T-ViT-14 (vanilla)&17.16\\
    ImageNet&LBG-T2T-ViT-14 (ours)&15.50\\
    ImageNet&DALBG-T2T-ViT-14 (ours)&\textbf{15.47}\\
    \cline{2-3}
    ImageNet&Swin-T (vanilla)&18.70\\
    ImageNet&LBG-Swin-T (ours)&16.81\\
    ImageNet&DALBG-Swin-T (ours)&\textbf{16.64}\\

    \hline
    \end{tabular}
    
    \label{tab:fixed2}
%
\end{table}

\begin{table*}[t]
\centering
\caption{Test errors, number of model parameters (in millions), and search costs (GPU days on a Tesla v100) on CIFAR-100 and CIFAR-10.  
    (DA)LBG-DARTS represents (DA)LBG applied to DARTS. Similar meanings hold for other notations in such a format.}
\resizebox{.99 \textwidth}{!}{
    \begin{tabular}{l||ccc|ccc}

    \hline
    \multicolumn{1}{l||}{} & \multicolumn{3}{c|}{CIFAR-100} & \multicolumn{3}{c}{CIFAR-10}\\ 
    \cline{2-7}
    Method & Error(\%)& Param(M)& Cost & Error(\%)& Param(M)& Cost\\
    \hline
\hline
DARTS \cite{liu2018darts}  & 20.58$\pm$0.44&3.4&1.5 &2.76$\pm$0.09&3.3&  1.5\\
LBG-DARTS (ours) &18.02$\pm$0.36 &3.6 & 1.7&\textbf{2.62$\pm$0.08}&3.5&1.6 \\
DALBG-DARTS (ours) &\textbf{17.97$\pm$0.43} &3.7 & 2.0&2.64$\pm$0.12&3.6&2.0 \\
\hline

PC-DARTS \cite{abs-1907-05737} &17.96$\pm$0.15&3.9&0.1& 2.57$\pm$0.07 & 3.6 & 0.1 \\
LBG-PCDARTS (ours) &16.21$\pm$0.19 &4.1 &0.3 &2.51$\pm$0.11&3.7&0.3 \\
DALBG-PCDARTS (ours) &\textbf{16.18$\pm$0.21} &4.2 &0.4 &\textbf{2.48$\pm$0.15}&3.8&0.4 \\
\hline
P-DARTS \cite{chen2019progressive}&17.49&3.6&0.3& 2.50&3.4&  0.3\\
LBG-PDARTS (ours) & 16.46$\pm$0.54& 3.7& 0.6&2.48$\pm$0.16&3.5& 0.5\\
DALBG-PDARTS (ours) & \textbf{16.39$\pm$0.48}& 3.9& 0.6&\textbf{2.47$\pm$0.19}&3.7& 0.6\\
\hline
    \end{tabular}}

    \label{tab:cifar100}
\end{table*}

\paragraph{Human-Designed GSCMs.} For experiments on CIFAR-10/100 and ImageNet datasets, we use ResNet \cite{he2016deep}, Swin-T \cite{liu2021swin}, and T2T-ViT \cite{yuan2021tokens} models as our base GSCM models in the conducted experiments. For consistency and a fair comparison, we apply $K=2$ latent subgroups to our four image classification datasets. To train our models, first we apply our proposed LBG training, where we use half of training images as the training set and the other half as the validation set, for 100 epochs with early stopping technique to obtain the optimal subgroups. Then, we use the obtained subgroups to fine-tune our GSCM models using the standard training settings with SGD optimizer for 200 epochs on the entire training examples.  The initial learning rate is set to 0.1 with momentum 0.9 and will be reduced using a cosine decay scheduler with the weight decay of 3e-4. The batch size for CIFAR-10 and CIFAR-100 is set to 128, while for ImageNet we use the batch size of 1024. The rest of hyperparameter settings follows as \cite{gururangan2020don}.
In all DALBG experiments we use $\lambda=0.1$. In this study, the Adam optimizer has been employed to train all models on the CelebA dataset, utilizing a learning rate of 5e-4, and implementing a batch size of 64. On the other hand, for the ISIC-18 experiments, we have set the learning rate to 1e-3, and the batch size to 32. The remaining settings are analogous to those of CIFAR10/100. 

\paragraph{GSCMs with Searchable Architectures.}
We apply LBG to various DARTS-based approaches: DARTS~\cite{liu2018darts}, P-DARTS~\cite{chen2019progressive}, and PC-DARTS~\cite{abs-1907-05737}. The search spaces of these methods are the combination of (dilated) separable convolutions with two different sizes of $3\times 3$ and $5\times 5$, max pooling with the size of $3\times 3$, average pooling with the size of $3\times 3$, identity, and zero operations. Each LBG experiment was repeated five times with different random seeds. The mean and standard deviation of classification errors obtained from the experiments are reported. 

In the architecture search stage, for  CIFAR-10 and CIFAR-100, the architecture of each group-specific classification model contains 5 cells -- reduced from 8 cells to 5 cells to match the parameter numbers of our baseline models -- and each cell consists of 7 nodes.  We use two group-specific sub-models (i.e., two subgroups $K=2$) in the search process with the initial channels of 16.  The search algorithm was based on SGD with a batch size of 64, the initial learning rate of 0.025 (reduced in later epochs using a cosine decay scheduler), epoch number of 50,  weight decay of 3e-4, and momentum of 0.9. The rest of hyperparameters mostly follow the original settings in DARTS, P-DARTS, and PC-DARTS. For a fair comparison, in all the DALBG-NAS experiments $\lambda = 0.1$.
For ISIC-18 and CelebA experiments, we utilize the same setting as described in the previous part. 


During architecture evaluation, each GSCM sub-model is formed by stacking 11 copies (reduced from 20 layers to align with the baselines' sizes) of the corresponding optimally searched cell for CIFAR-10 and CIFAR-100 experiments. The initial channel number is set to 36. We train the networks with a batch size of 96 and 600 epochs on a single Tesla V100 GPU.  
For evaluation of ImageNet, we use the searched architectures on CIFAR-10 and we stack 8 copies (similarly reduced from 14 layers to match the baselines' sizes) of obtained cells are stacked into each GSCM larger network, which was trained using four Tesla V100 GPUs on the 1.2M training images, with the batch size of 1024 and initial channel number of 48 for 250 epochs. Finally, for ISIC-18 and CelebA architecture evaluation we utilize the same setting as the one described for fixed human-designed GSCMs.

\subsection{Results}
\label{sec:4.3}


First, we evaluate and compare the fairness of our proposed methods with the our baselines on CelebA dataset. In line with the methodology of \cite{wang2022fairness}, we use "attractive" as the binary class label for prediction, and as bias-sensitive attribute, we consider "gender" (male and female) in relation to the predicted labels. 
For evaluation we use Demographic Parity (DP) and Difference in Equalized Odds (DEO) metrices similar to \cite{wang2022fairness}.
The results of our experiments, as shown in Table~\ref{tab:fair}, demonstrate that our proposed methods can improve accuracy while simultaneously mitigating unfair decision-making on minority groups. This is achieved through the use of group-specific models, which are trained on individual groups. We can also observe that DALBG improves accuracy more than LBG, but LBG achieves better fairness results on the DEO metric. This could be due to the fact that domain adaptation incorporated in DALBG may perpetuate or even amplify any existing biases present in the source domain, which may not be fully removed if the target domain is significantly different.

Table~\ref{tab:isic-fair} demonstrates the results of fairness experiments on the ISIC-18 dataset where gender is the sensitive attribute and we use Statistical Parity Difference (SPD), Equal Opportunity Difference (EOD), and Average Odds Difference (AOD) as metrics to evaluate a model's fairness. This table shows that our method not only boosts accuracy performance, but also improves fairness by effectively mitigating bias in both fixed human-designed neural networks and NAS. This performance improvement is attributable to our group-aware approach, which effectively groups similar samples with respect to unprotected sensitive attributes. This proves the advantage of our methods in addressing imbalanced attributes in the data.

Furthermore, in Table~\ref{tab:fixed2}, we compare our proposed method with ResNet, Vision Transformers (Swin-T and T2T-ViT) ,and our MoE baselines (i.e., MGE-CNN and DeepMOE). The results in this table verify that our proposed method performs better than the baselines on all CIFAR-10, CIFAR-100, and ImageNet datasets considerably. This empirically verifies our claim that (DA)LBG  reduces the overfitting risk found in MoE methods  since the group assignments are learned by minimizing the validation loss during a multi-stage optimization.

Table~\ref{tab:cifar100} shows the comparison of our proposed methods and the existing works, which includes the classification errors with error bars, the number of model parameters, and search costs on CIFAR-10 and CIFAR-100 test sets. By comparing different methods, we make the following observation. Applying (DA)LBG to different NAS methods, including DARTS, P-DARTS, and PC-DARTS, the classification errors of these methods are greatly reduced. For instance, the original error of DARTS on CIFAR-100 is 20.58\%; when DALBG is applied, this error is significantly reduced to 17.97\%. As another example, after applying LBG to PC-DARTS and P-DARTS, the errors of CIFAR-100 experiments are decreased from 17.96\% to 16.21\% and 17.49\% to 16.46\%, respectively. Similarly for CIFAR-10, utilizing (DA)LBG in DARTS-based methods manages to reduce the errors and overfittings.
These results strongly indicate the broad effectiveness of our framework in searching better neural architectures.

\begin{table}[ht]
\centering
\vspace{-.5cm}
\caption{Results of ImageNet with gradient-based NAS methods. Notations are the same as those in Table~\ref{tab:cifar100}. }
\vspace{.2cm}
\resizebox{.99\columnwidth}{!}{
    \centering
    \begin{tabular}{l|ccc}
    \toprule
  & Top-1  &Top-5 &Param  \\
         Method& Error (\%) & Error (\%)&(M) \\
    \midrule
       \hline
       DARTS-CIFAR10 \cite{liu2018darts}  & 26.7 &8.7&4.7\\
        DALBG-DARTS-CIFAR10 (ours) &\textbf{24.9} &8.1 &4.9  \\
        \hline
          P-DARTS (CIFAR10) \cite{chen2019progressive}&24.4 &7.4&4.9\\
        DALBG-PDARTS-CIFAR10 (ours) &\textbf{23.9} &6.9 &5.0 \\
        \hline
            PC-DARTS-CIFAR10 \cite{abs-1907-05737} &  24.8 &7.3&5.3\\
              DALBG-PCDARTS-CIFAR10 (ours)&\textbf{23.1} &6.3 & 5.7\\
        \bottomrule
    \end{tabular}
    }
    
    \label{tab:imagenet}
\end{table}

In Table~\ref{tab:imagenet}, we compare different methods on ImageNet, in terms of top-1 and top-5 errors on the test set and number of model parameters, where the search costs are the same as the ones reported in Table~\ref{tab:cifar100}. 
In these experiments, the architectures are searched on CIFAR-10 and evaluated on ImageNet similar to original DARTS \cite{liu2018darts}. DALBG-DARTS-CIFAR10 denotes that DALBG is applied to DARTS and performs search on CIFAR-10. 
Similar meanings hold for other notations in such a format. 
The observations made from these results are consistent with those made from Table~\ref{tab:cifar100}. The architectures searched using our methods are consistently better than those searched by corresponding baselines. For example, DALBG-DARTS-CIFAR10 achieves 1.8\% lower top-1 error than DARTS-CIFAR10. To the best of our knowledge, DALBG-PCDARTS-CIFAR10 is the new SOTA on mobile setting of Imagenet.




\subsection{Ablation Studies}

In this section, we conduct ablation studies to analyze the impact of individual components in our proposed frameworks.

\paragraph{Ablation on tradeoff parameter $\lambda$:}
We study the effectiveness of tradeoff parameter $\lambda$ in Eq. (\ref{DA}) on accuracy and fairness. We apply DALBG on CelebA dataset with two searchable GSCMs (i.e., $K=2$) with the same setting as described in Section \ref{sec:4.2}.
In Table \ref{tab:ab1}, we illustrate how the accuracy and fairness of DALBG on the test sets of CelebA are affected by increasing the tradeoff parameter $\lambda$. It can be observed that increasing $\lambda$ from 0 to 0.1 leads to a decrease in fairness but an increase in accuracy, as a result of the MMD loss feedback. However, continuing to increase $\lambda$ leads to a drop in accuracy as well. This is because placing too much emphasis on domain shift can result in less focus on in-domain performance ability.
\vspace{-.3cm}
\begin{center}
\begin{table}[h]
\centering
\caption{Ablation results on tradeoff parameter $\lambda$.}
    \centering
    \begin{tabular}{l|cc}
  \hline
  Methods &  Error (\%)  &DEO  \\

       \hline
       \hline
       LBG-DARTS  with $\lambda = 0$ &  15.91 & \textbf{0.0535} \\
       DALBG-DARTS  with $\lambda = 0.01$ &  15.73 & 0.0754 \\
       DALBG-DARTS  with $\lambda = 0.1$ &  \textbf{15.22} & 0.0891 \\
       DALBG-DARTS  with $\lambda = 1$ &   15.38 & 0.0917 \\
\hline
    \end{tabular}
    \label{tab:ab1}  
\end{table}
\end{center}
\paragraph{Ablation on number of subgroups:}
Next, we examine how different numbers of GSCM models with different number of subgroups $K\in \{1,2,3,4\}$ in Eq. (\ref{NASOP}) impact both accuracy and fairness performances. We apply (DA)LBG to DARTS. Table \ref{tab:ab2} indicates that for CelebA larger number of subgroups can decrease the classification error and improve the fairness. However, in our experiments  number of subgroups $K=3$ and $K=4$ seem to achieve on par results, while  $K=3$ is more computationally efficient. The improved performance with a larger number of subgroups can be due to the fact that, in real life, many unprotected attributes may not be considered, but their combinations can still be used as proxies and affect decision-making processes. Thus, depending on the data and task, we can choose the most suitable number of subgroups, which can be different in various scenarios. Also, additional experiments and comparisons of (DA)LBG with bagging-based model ensemble can be found in the Supplements.
\vspace{-.3cm}
\begin{center}
\begin{table}[h]
\centering
\caption{Ablation results on number of subgroups.}
\resizebox{.99\columnwidth}{!}{
    \centering
    \begin{tabular}{l|cc}
  \hline
  Methods &  Error (\%)  &DEO  \\

       \hline
       \hline
       LBG-ResNet18  with $K=1$ &  17.59 & 0.5427 \\
       LBG-ResNet18  with $K=2$ &  17.02 & 0.0596 \\
       LBG-ResNet18  with $K=3$ &  16.88 & 0.0541 \\
       LBG-ResNet18  with $K=4$ &  \textbf{16.85} & \textbf{0.0533} \\
\hline
    \end{tabular}
    }
    \label{tab:ab2}  
\end{table}
\end{center}
\paragraph{Ablation on different domain adaptation techniques:}
In this study, we aim to investigate the efficacy of different distance measuring approaches, namely Maximum Mean Discrepancy (MMD), Correlation Alignment (CORAL), Kullback-Leibler (KL) divergence, and Contrastive Domain Discrepancy (CDD), by incorporating them into our framework. We conduct our experiments on DARTS with a similar experimental setup to Table \ref{tab:fair}. The results, presented in Table \ref{tab:ab3}, indicate that MMD loss is the most effective approach in achieving both high accuracy and fairness compared to the other three methods. The superior performance of MMD in our framework can be attributed to its non-parametric nature and ability to capture non-linear relationships due to its kernel-based approach. In contrast, KL divergence relies on the assumption that both distributions are well-defined probability distributions and, along with CDD, may struggle to capture non-linear relationships in the data. Furthermore, while CORAL aligns the second-order statistics (i.e., covariance matrices) of the feature distributions, MMD maps the data into a Reproducing Kernel Hilbert Space (RKHS) using kernel functions. This capability enables MMD to capture more intricate relationships between data points, potentially resulting in improved performance within our framework. However, it is worth noting that the effectiveness of a domain adaptation technique may vary depending on the specific task and the degree of domain shift between the source and target domains. Thus, the choice of an appropriate technique should be based on the unique characteristics of the data and the task at hand.
\vspace{-.3cm}
\begin{center}
\begin{table}[h]
\centering
\caption{Ablation results on different domain adaptation techniques.}
\resizebox{.99\columnwidth}{!}{
    \centering
    \begin{tabular}{l|cc}
  \hline
  Methods &  Error (\%)  &DEO  \\

       \hline
       \hline
       DALBG-DARTS-MMD  &  \textbf{15.22} & \textbf{0.0891}\\
       DALBG-DARTS-CORAL &  15.71 & 0.1137 \\
       DALBG-DARTS-KL  &  15.36 & 0.0945 \\
       DALBG-DARTS-CDD   &  15.80 & 0.1142 \\
\hline
    \end{tabular}
    }
    \label{tab:ab3}  
\end{table}
\end{center}
\section{Conclusions and Discussion}
\label{sec:5}
In this paper, we propose a novel MLO approach, called \textit{Learning by Grouping} (LBG),  drawing from humans' grouping-driven methodology of solving problems. Our approach learns to group a diverse set of problems into distinct subgroups where problems in the same subgroups are similar; a group-specific solution is developed to solve problems in the same subgroups. We formulate our LBG as a multi-level optimization problem which is solved end-to-end. An efficient gradient-based optimization algorithm is developed to solve the LBG problem. We further incorporate  domain adaptation in our framework to reduce the risk of overfitting. In our experiments on various datasets, we demonstrate that the proposed framework not only helps to mitigate overfitting and improve fairness, but also consistently outperforms baseline methods.  The main limitation of LBG is that it cannot be applied to non-differentiable NAS approaches up to a point. In our future works, we will extend learning by grouping to reinforcement learning and evolutionary algorithms. 


\bibliography{refs}
\bibliographystyle{icml2023}

\newpage
\appendix
\onecolumn
\section{Additional Experiments}


In this section, we apply LBG with fixed human-designed architectures to language understanding tasks. 

\subsection{Datasets}
 We conducted experiments on the various tasks of the General Language Understanding Evaluation (GLUE) benchmark \cite{wang-etal-2018-glue}. GLUE contains nine tasks, which are two single-sentence tasks (CoLA and SST-2), three similarity and paraphrase tasks (MRPC, STS-B, and QQP), and four inference tasks (MNLI, QNLI, RTE, WNLI). We test the performance of LBG in language understanding by submitting our inference results to the GLUE evaluation server. GLUE offers training and development data splits, that are used as training and validation data. For the  test dataset, and GLUE organisers provide a submission server that reports the performance on the private held out test dataset.

\begin{table*}[ht]
\centering
\caption{Comparison of BERT-based and RoBERTa-based experiments on GLUE sets. LBG-BERT and LBG-RoBERTa results on the set are the medians of 5 runs.}
\vspace{.2cm}
\resizebox{ \textwidth}{!}{
     \centering

    \begin{tabular}{l||c|c||c|c}
    \hline
    Corpus & \textbf{BERT}&\textbf{LBG-BERT}&\textbf{RoBERTa}&\textbf{LBG-RoBERTa}\\
    \hline
    CoLA (Matthews Corr.)&60.5&\textbf{62.8}&68.0&\textbf{69.5}\\
    SST-2 (Accuracy)&94.9&\textbf{96.5}&96.4&\textbf{96.8}\\
    MRPC (Accuracy/F1)&85.4/89.3&\textbf{86.2/89.5}&\textbf{90.9}/92.3&90.2/\textbf{92.4}\\
    STS-B (Pearson/Spearman Corr.)&87.6/86.5&\textbf{88.4/87.9}&92.4/92.0&\textbf{92.5/92.3}\\
    QQP (Accuracy/F1)&89.3/72.1&\textbf{89.6/72.3}&92.2/-&\textbf{92.6/77.0}\\
    MNLI (Matched/Mismatched Accuracy)&\textbf{86.7/85.9}&86.5/\textbf{85.9}&90.2/90.2&\textbf{91.1/91.1}\\
    QNLI (Accuracy)&92.7&\textbf{93.5}&94.7&\textbf{94.9}\\
    RTE (Accuracy)&70.1&\textbf{72.4}&86.6&\textbf{86.7}\\
    WNLI (Accuracy)&65.1&\textbf{66.3}&\textbf{91.3}&86.3\\
    \hline
    \end{tabular}
    }
    \label{tab:fixed1}
\end{table*}


\subsection{Experimental Settings}
We examine our proposed method by conducting varied experiments on several different tasks and datasets. We compare LBG on language understanding tasks with fixed architectures using BERT \cite{devlin-etal-2019-bert} and RoBERTa \cite{liu2019roberta}. BERT \cite{devlin-etal-2019-bert} and RoBERTa \cite{liu2019roberta} initialize the Transformer encoder with pre-trained BERT and RoBERTa, respectively, with the intentions of masked language modeling and next sentence prediction. Then, they utilize the pre-trained encoder and a classification head to build a text classification model. This text classification model latter will be fine-tuned on a target classification task.

To examine our method in language understanding, we employ BERT and RoBERTa as the group-specific sub-models with $K=4$ latent subgroups on the GLUE tasks. LBG-BERT and LBG-RoBERTa are optimized using Adam optimizer \cite{paszke2017automatic}. The maximum length of text was set to 512 tokens. Our hyperparameter settings for BERT and RoBERTa experiments are the same as in \cite{gururangan2020don}. Each GLUE task has a different batch size, learning rate, and number of epochs, where they are within the batch sizes $\in \{16,32\}$, learning rates $\in \{1e^{-5}, 2e^{-5}, 3e^{-5}, 4e^{-5}\}$, and number of epochs $\in \{3,4,5,6,10\}$. More detailed hyperparameter settings are exhibited in the Appendix- Supplementary Materials.

\subsection{Results}

Table~\ref{tab:fixed1} demonstrates the comparison of our methods with BERT and RoBERTa methods on nine different GLUE tasks. It is shown in this table that LBG can efficiently enhance the performance of existing base models in various language understanding tasks. In most of the tasks, LBG-BERT and LBG-RoBERTa outperform BERT and RoBERTa, respectively. In MNLI and MRPC, the results of our methods are on par with the baselines, while RoBERTa achieves a slightly better result than our methods on the WNLI task. 

\end{document}